\pdfoutput=1
\documentclass[]{dataflow}

\usepackage{latexsym}
\usepackage{textcomp}
\usepackage{amsmath}
\usepackage{array}
\usepackage{makecell}
\usepackage{float}
\usepackage{flafter}
\usepackage[ruled,vlined]{algorithm2e}
\usepackage{comment}
\usepackage{pifont}
\usepackage{fontawesome5}
\usepackage{url}
\graphicspath{{./}}
\makeatletter
\def\input@path{{./}}
\makeatother

\usepackage[table]{xcolor}
\newcommand{\model}{OmniEdu\xspace}
\usepackage[most]{tcolorbox}
\definecolor{model_type}{RGB}{224,225,221}

\definecolor{rankone}{HTML}{B43735}
\definecolor{ranktwo}{HTML}{C85F1A}
\definecolor{rankthree}{HTML}{C5A51F}
\newcommand{\rankOne}[1]{\textcolor{rankone}{\textbf{#1}}}
\newcommand{\rankTwo}[1]{\textcolor{ranktwo}{\textbf{#1}}}
\newcommand{\rankThree}[1]{\textcolor{rankthree}{\textbf{#1}}}

\title{OmniEdu: Open Foundation Models for Learning and Teaching}

\author[*,1,3]{Hao Liang}
\author[*,1]{Qihan Lin}
\author[1]{Meiyi Qiang}
\author[2]{Linzhuang Sun}
\author[1]{Hengyi Feng}
\author[2]{Mingrui Chen}
\author[2]{Sizhe Qiu}
\author[\ddagger,1,3]{Wentao Zhang}

\affiliation[1]{Peking University}
\affiliation[2]{University of the Chinese Academy of Sciences}
\affiliation[3]{Zhongguancun Academy}

\contribution[*]{Equal contribution}
\contribution[\ddagger]{Corresponding author}

\def\emailicon{\raisebox{-1.5pt}{\includegraphics[height=1.05em]{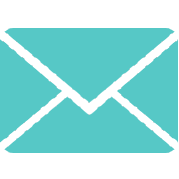}}}
\def\githubicon{\raisebox{-1.5pt}{\includegraphics[height=1.05em]{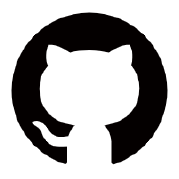}}}
\def\huggingfaceicon{\raisebox{-1.5pt}{\includegraphics[height=1.05em]{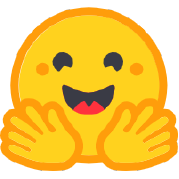}}}
\def\websiteicon{\raisebox{-0.5pt}{\faGlobe}}

\newcommand{\sourcelink}{https://github.com/haolpku/Omni-Edu}
\newcommand{\websitelink}{https://github.com/haolpku/Omni-Edu}
\checkdata[\emailicon\hspace{0.3em} Email]{\email{hao.liang@stu.pku.edu.cn}}
\checkdata[\websiteicon\hspace{0.3em} Project Website]{\url{\websitelink}}
\checkdata[\githubicon\hspace{0.3em} Source Code and Dataset]{\url{\sourcelink}}
\checkdata[\huggingfaceicon\hspace{0.3em} Training Dataset]{\url{https://huggingface.co/datasets/lhpku20010120/Omni-Edu}}
\checkdata[\huggingfaceicon\hspace{0.3em} OmniEdu-4B]{\url{https://huggingface.co/lhpku20010120/Omni-Edu-4B}}
\checkdata[\huggingfaceicon\hspace{0.3em} OmniEdu-9B]{\url{https://huggingface.co/lhpku20010120/Omni-Edu-9B}}
\checkdata[\huggingfaceicon\hspace{0.3em} OmniEdu-27B]{\url{https://huggingface.co/lhpku20010120/Omni-Edu-27B}}

\abstract{%
Educational foundation models must do more than produce correct answers: they must understand where a problem sits in a curriculum, diagnose why a learner is struggling, and choose an appropriate instructional response. Existing educational language models often specialize in either subject problem solving or tutoring, while their training mixtures are commonly organized by source or task and do not explicitly balance these capabilities. We present \textsc{OmniEdu}, an open family of foundation models for K--12 learning and teaching, trained with a capability-oriented instruction-tuning corpus. The corpus combines more than 100 educational resources and general instruction sources and organizes supervision around four complementary capabilities: subject competence, curriculum grounding, diagnostic reasoning, and pedagogical action and scaffolding. A multi-stage pipeline performs deterministic cleaning, semantic auditing and rewriting, task-specific quality scoring, token-budgeted diversity selection, and pedagogical instruction assignment, yielding 69{,}999 examples and 15.96M supervised response tokens, including 60{,}951 education-specific examples. We fine-tune 4B, 9B, and 27B models and evaluate them on curriculum-grounding, K--12 problem-solving, and pedagogical-tutoring benchmarks, with auxiliary tests of general capability. Across model scales, education-oriented tuning consistently improves all three educational capability groups. In particular, \textsc{OmniEdu}-27B reaches 63.12\% EM / 76.69\% F1 on K12-Bench, 85.89\% on MathFish, 86.95\% on EDUMATH, 78.74\% on MathTutorBench's Scaffold setting, and the best Teaching average of 3.02 on LongTutor among the evaluated models. These results show that carefully curated, capability-balanced supervision can turn a general language model into a stronger educational system that not only solves problems, but also understands curriculum structure and supports effective teaching interactions.}

\begin{document}
\maketitle

\renewcommand{\thefootnote}{\fnsymbol{footnote}}
\setcounter{footnote}{0}
\renewcommand{\thefootnote}{\arabic{footnote}}
\pagestyle{fancy}
\fancyhf{}
\fancyhead[L]{OpenDCAI Technical Report}
\fancyhead[R]{\thepage}

\newpage
\tableofcontents
\clearpage

\section{Introduction}
\label{sec:introduction}

Large language models are increasingly used in education, but educational usefulness is not captured by answer accuracy alone. A capable learning and teaching assistant must solve a student's problem, connect it to the appropriate knowledge point and prerequisite structure, identify the misconception behind an incorrect attempt, and select an intervention that advances learning. In other words, an educational model must coordinate \emph{what to teach}, \emph{where the learner is}, and \emph{how to respond}. Treating education as ordinary question answering therefore leaves out the structure that makes tutoring effective.

Recent educational models have made progress in individual parts of this problem. Some emphasize subject competence and examination-style problem solving; others target tutoring dialogue, Socratic questioning, or curriculum alignment~\cite{educhat,muduollm,confucius3math,learnlm}. However, these capabilities are often developed in separate systems or evaluated in isolation. A model may obtain the right answer while failing to locate the relevant curriculum concept, explain an error, or provide a scaffold that preserves the learner's opportunity to reason. The field consequently lacks an open model family whose training objective and evaluation protocol are explicitly organized around the full learning--teaching loop.

We argue that this gap is partly a data-design problem. Educational instruction data are heterogeneous: a short answer, a curriculum relation, a misconception diagnosis, and a tutoring exchange supervise different behaviors. Mixing them by source or subject alone can obscure the intended response policy and can overrepresent easy or redundant examples. The central design question is therefore not simply how to collect more educational data, but how to construct a training mixture in which each example has a clear capability target and a pedagogically appropriate response behavior.

To address this question, we present \textsc{OmniEdu}, an open family of foundation models for K--12 learning and teaching. We organize the education-specific supervision around four complementary capabilities. \emph{Subject competence} covers solving K--12 problems and explaining answers across mathematics, science, reading, and writing. \emph{Curriculum grounding} captures knowledge-point alignment, grade level, difficulty, prerequisite relations, and curriculum localization. \emph{Diagnostic reasoning} requires the model to identify errors and misconceptions and infer missing prerequisites from a learner's work. \emph{Pedagogical action and scaffolding} teaches the model to select and execute an appropriate intervention, such as a hint, a question, a prerequisite review, or a direct explanation. This taxonomy makes the intended educational behavior explicit and provides a common language for both data construction and evaluation.

We implement the taxonomy in a data-centric post-training pipeline. Starting from more than 100 datasets and educational resources, we combine deterministic cleaning and decontamination with LLM-assisted semantic auditing, task-specific quality scoring, and diversity-aware selection. The pipeline reduces an initial education-specific pool of approximately 1.34M examples to 60{,}951 high-quality examples containing about 12.0M supervised response tokens. We then add 9{,}048 general-purpose instruction examples and attach one of 20 task-specific pedagogical system instructions to each selected example. The resulting mixture contains 69{,}999 examples and 15.96M supervised response tokens, with provenance and audit metadata retained throughout. This construction separates the content of an example from the response behavior it is intended to teach, allowing the same model to support both answer-oriented and scaffold-oriented interactions.

We fine-tune 4B, 9B, and 27B models and evaluate them along three educational dimensions: curriculum grounding, K--12 problem solving, and pedagogical tutoring. The evaluation spans curriculum-structure benchmarks (K12-Bench, MathFish, and EDUMATH), authentic school-level problem-solving benchmarks (GAOKAO-Bench, EXAMS-V, and MDK12-Bench), and tutoring benchmarks that test scaffolding, feedback, adaptive explanation, and longitudinal student modeling (MathTutorBench, TutorBench, and LongTutor). We additionally measure general instruction following, scientific reasoning, and multimodal understanding to assess the broader effects of educational specialization.

The results support the capability-oriented design. Education-oriented tuning improves every model scale across the three educational dimensions. \textsc{OmniEdu}-27B obtains 63.12\% EM and 76.69\% F1 on K12-Bench, 85.89\% accuracy on MathFish, and 86.95\% MaC on EDUMATH. On tutoring, it reaches 78.74\% Scaffold win rate on MathTutorBench and a Teaching average of 3.02 on LongTutor, while remaining competitive with substantially larger proprietary systems on several K--12 problem-solving evaluations. The gains are therefore not limited to producing more correct answers: they also appear in curriculum reasoning, error-sensitive teaching, and the use of student history. Auxiliary general-capability evaluations provide an additional check on the cost of specialization.

Our contributions are:
\begin{itemize}
    \item We introduce \textsc{OmniEdu}, an open family of K--12 foundation models organized around four capabilities that connect subject knowledge, curriculum structure, learner diagnosis, and pedagogical action.
    \item We develop a reproducible capability-oriented data construction pipeline that combines semantic quality control, task-specific filtering, token-budgeted diversity selection, and explicit pedagogical instructions, producing a 69{,}999-example training mixture with 15.96M supervised response tokens.
    \item We provide a cross-capability evaluation showing consistent gains across curriculum grounding, K--12 problem solving, and pedagogical tutoring at 4B, 9B, and 27B scales, together with auxiliary tests of general model capability. The results establish data composition and pedagogical behavior as central design axes for open educational foundation models.
    \item We provide public project and dataset entry points for reproducibility through the \href{https://github.com/haolpku/Omni-Edu}{project repository} and the \href{https://huggingface.co/datasets/lhpku20010120/Omni-Edu}{Hugging Face dataset page}.
\end{itemize}


\section{Related Work}
\label{sec:related_work}

\subsection{Educational Foundation Models and AI Tutors}

Recent work has increasingly adapted general-purpose LLMs to educational scenarios.
Among open educational models, EduChat~\cite{educhat} targets a broad range of learning interactions, including question answering, essay assessment, Socratic teaching, and emotional support, while MuduoLLM~\cite{muduollm} is aligned with the Chinese K--12 curriculum and supports tasks such as problem solving, guided question answering, question generation, and lesson planning.
Other models focus more narrowly on particular educational capabilities. 
For example, Confucius3-Math~\cite{confucius3math} specializes in mathematical reasoning for Chinese K--12 education.
These models demonstrate the value of domain-specific adaptation, but differ substantially in their coverage of subject knowledge, curriculum understanding, and pedagogical interaction.

In parallel, frontier general-purpose models and AI tutoring systems increasingly incorporate explicit pedagogical behaviors.
LearnLM~\cite{learnlm} formulates educational adaptation as \emph{pedagogical instruction following}, allowing teaching behaviors to be conditioned on the learning scenario during post-training.
Product systems such as ChatGPT Study Mode~\cite{study_mode} and Khanmigo~\cite{khanmigo} similarly emphasize guided reasoning, scaffolding, and learner engagement rather than simply providing answers, while recent teacher-facing systems~\cite{anthropic} further integrate curriculum resources and structured teaching workflows.
However, many of these frontier systems remain closed or rely heavily on product-level orchestration, making their underlying educational capabilities difficult to reproduce and extend.
\model complements these efforts by providing an open model family that jointly targets both learning- and teaching-oriented capabilities.

\subsection{Educational Data and Data-Centric Post-Training}

A growing body of work has shown that instruction-tuning performance depends strongly on the quality and composition of training data rather than raw data scale alone.
LIMA~\cite{lima} demonstrates that carefully curated supervision can induce strong instruction-following behavior with only a small number of examples.
AlpaGasus~\cite{alpagasus} filters low-quality instruction data with a strong LLM, while Deita~\cite{deita} explicitly considers data quality, complexity, and diversity for efficient instruction tuning.
LESS~\cite{less} further studies targeted data selection by identifying examples that are particularly influential for desired downstream capabilities.
Together, these works motivate a data-centric view of post-training in which quality, diversity, and capability relevance are central considerations.

Educational data introduces additional structure beyond generic instruction tuning.
Instruction Tuning with Human Curriculum~\cite{human_curriculum} organizes synthetic instructions according to educational subject hierarchies and difficulty progression.
K12-KGraph~\cite{k12kgraph} derives curriculum-structured supervision from a knowledge graph extracted from official K--12 textbooks, while EDUMATH~\cite{edumath} constructs targeted supervision for standards-aligned math word-problem generation.
These studies show the benefit of incorporating educational structure into training data, but typically focus on a particular task, subject, or capability.
\model extends these ideas to a unified, capability-oriented educational post-training mixture.

\subsection{Evaluation of Educational LLMs}

Existing benchmarks for educational LLMs cover a broad spectrum of capabilities and can be roughly grouped into learner competence, curriculum understanding, and pedagogical capability.
Learner-oriented benchmarks primarily assess subject knowledge, problem solving, and reasoning through school examinations and academic tasks, spanning text-only, multilingual, and increasingly multimodal settings~\cite{scienceqa,gaokaobench,eeval,m3exam,cmmu,k12vista,examsv,mdk12,k12edubench}.
Recent work also moves beyond final-answer accuracy toward process-level reasoning and more fine-grained assessment~\cite{gaokaoeval,k12pebench}.
Curriculum-oriented benchmarks examine whether models understand how educational content is organized, including alignment with standards, grade levels, knowledge points, prerequisite structures, and the generation of curriculum-aligned content~\cite{mathfish,k12kgraph,cjeval,eduadapt,knowshiftqa,edumath}.
Pedagogy-oriented benchmarks evaluate how models teach rather than merely what they know, covering explanation, scaffolding, feedback, error diagnosis, adaptive tutoring, multimodal interaction, and long-term student modeling~\cite{mathtutorbench,tutorbench,longtutor,mrbench,mmtutorbench,kmpbench,safetutors}.
This progression reflects a broader shift in educational AI evaluation from measuring whether a model can solve a problem to whether it can also understand what should be taught and how to teach it.


\section{Data Preparation}
\label{sec:data}

\begin{figure*}[htbp]
    \centering 
    \makebox[\textwidth]{\includegraphics[width=1.00\textwidth]{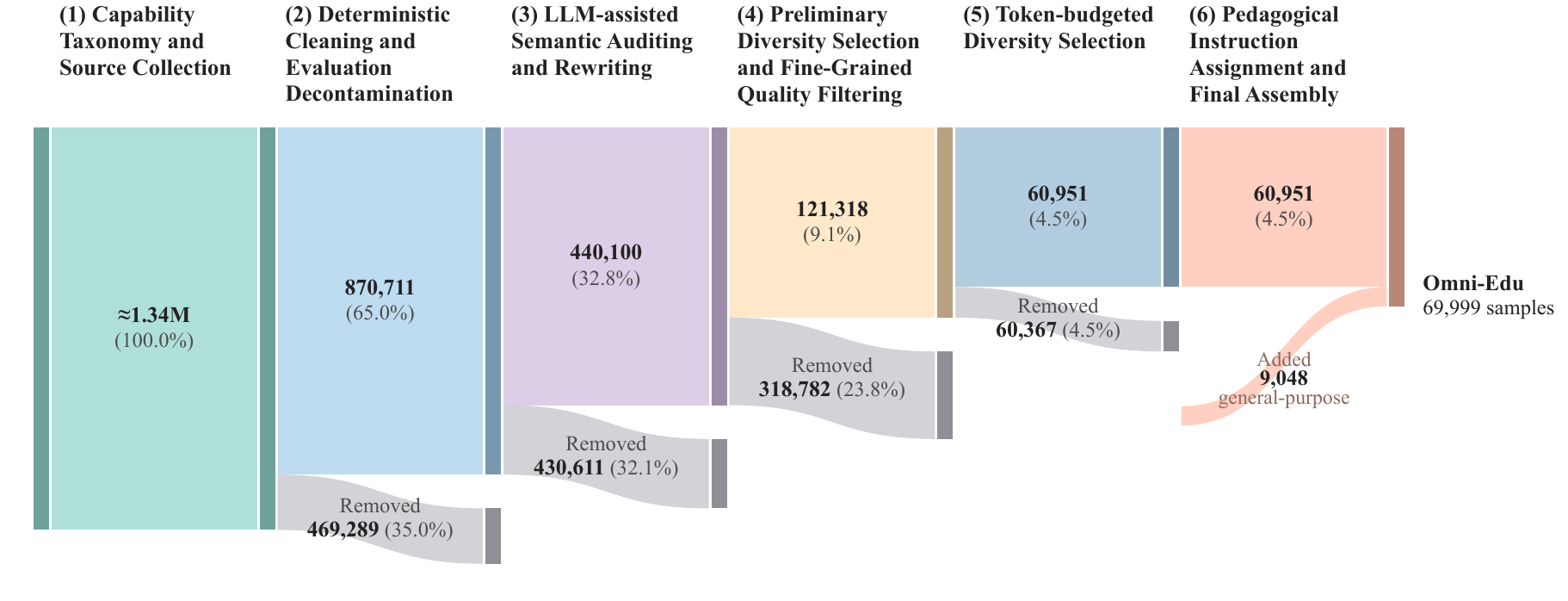}}
    \caption{Overview of the \textsc{OmniEdu} data preparation pipeline. The education-specific pool is progressively curated from approximately 1.34M candidates to 60,951 high-quality examples through six stages. Together with 9,048 general-purpose instruction examples, the final corpus contains 69,999 examples. Gray branches indicate filtered-out examples.}
    \label{fig:data_pipeline}
\end{figure*}

We construct a large-scale instruction-tuning corpus for K–12 education by integrating heterogeneous data sources and curating them according to the capabilities required of an educational language model. 
Our data preparation pipeline consists of six stages: (1) capability taxonomy and source collection, (2) deterministic cleaning and evaluation decontamination, (3) LLM-assisted semantic auditing and rewriting, (4) preliminary diversity selection and fine-grained quality filtering, (5) token-budgeted diversity selection, and (6) pedagogical instruction assignment and final assembly. 
Example-level provenance and audit metadata are retained throughout the pipeline.

We collect data from more than 100 datasets and educational resources. 
Before filtering, the education-specific pool contains approximately 1.34M examples. 
The following subsections describe how this pool is progressively curated into the final training mixture.

\subsection{Capability Taxonomy and Source Collection}
Our training corpus consists of two complementary components: \textbf{general-purpose instruction data} and \textbf{education-specific data}. 
The former contains 9,048 examples drawn from DataFlow-Instruct-10K (7,431), the Tulu-3-SFT-mixture (1,495), and MathV360K (122), in order to preserve the general instruction following, multilingual coverage, refusal behaviour, and diagram-based reasoning.
The latter constitutes the main focus of our data preparation pipeline and targets the knowledge, reasoning, and pedagogical capabilities required in K--12 educational scenarios.

Rather than organizing the education-specific data solely by subject or source, we define four capability categories according to the behaviors to be learned. 
\textit{Subject competence} covers solving K--12 problems and producing correct answers and explanations across mathematics, science, reading comprehension, and writing-related tasks. 
\textit{Curriculum grounding} captures curriculum structure, including grade-level alignment, knowledge-point identification, prerequisite relationships, difficulty, and the placement of problems within a curriculum.
\textit{Diagnostic reasoning} focuses on identifying errors and misconceptions in student solutions and inferring missing prerequisite knowledge. 
\textit{Pedagogical action and scaffolding} covers both selecting an appropriate instructional action (such as asking a question, providing a hint, revisiting a prerequisite concept, or giving a direct explanation) and carrying it out appropriately. 
A complete source inventory, licensing information, and source-to-category mapping are retained as release metadata.

Each example is assigned to a single primary capability according to its main supervision objective; other relevant properties are retained as secondary metadata. 
For subsequent mixture construction, we further partition examples within each capability into \textbf{fine-grained task buckets} according to factors such as task form, modality, and supervision type.

\subsection{Deterministic Cleaning and Evaluation Decontamination}

We first standardize the heterogeneous sources into a unified representation, remove exact duplicates, and discard examples with missing or malformed required fields.
For examples whose answers can be deterministically verified, we additionally check the consistency between the provided answer and the corresponding reference or structured annotation.
For mixed-domain sources, we retain only examples relevant to K--12 education and filter out unrelated content, such as finance or general encyclopedic knowledge.
For multimodal examples, we additionally verify that referenced images are accessible and correctly aligned with the textual input. 
Source and license information is retained for provenance tracking, and data with unclear usage conditions are excluded.
To prevent evaluation contamination, we remove any example that overlaps with our final evaluation set. 

After this stage, 870,711 education-specific examples remain and are passed to semantic auditing.

\subsection{LLM-assisted Semantic Auditing and Rewriting}
\label{sec:llm-audit}

We use Qwen3.5-122B-A10B-FP8 to perform semantic quality control that cannot be reliably handled by deterministic rules. 
Each example is evaluated using a task-specific auditing prompt and assigned a 0--100 usability score together with one of three actions: \textit{keep}, \textit{rewrite}, or \textit{remove}. 
Examples scoring 85--100 are retained, those scoring 50--84 are sent for repair, and those scoring below 50 are discarded.

The auditing criteria are adapted to the supervision target. 
For subject competence, the auditor focuses on answer correctness, consistency between the answer and explanation, and grounding in the given problem or passage. 
Curriculum examples are checked for valid curriculum relations and sufficiently specific knowledge-point alignment. 
Diagnostic data are assessed according to whether the identified error is supported by the student's work and whether the diagnostic explanation and corrective response are consistent with that error. 
Pedagogical examples are additionally evaluated for instructional relevance, coherence, scaffolding quality, and premature leakage of the final answer.

For examples labeled \textit{rewrite}, we keep the original input fixed and only repair the defective supervision. 
The rewritten example is then audited again using the same criteria and is retained only if it is classified as \textit{keep}. 
The full auditing prompts and criteria are retained with the release metadata.

After this stage, 440,100 education-specific examples remain and are passed to fine-grained quality filtering.

\begin{figure*}[htbp]
    \centering 
    \makebox[\textwidth]{\includegraphics[width=1.00\textwidth]{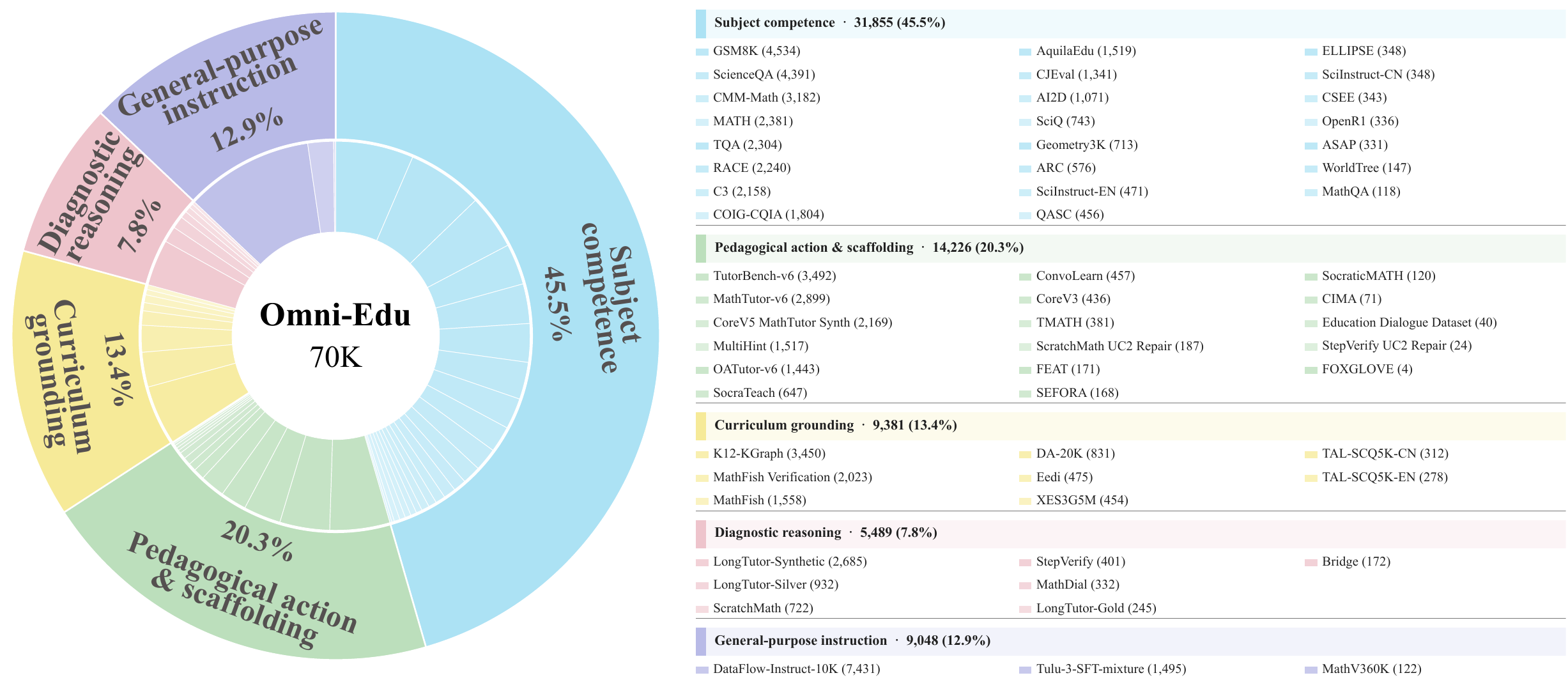}}
    \caption{Composition of the final \textsc{OmniEdu} (69{,}999 examples / 15.96M supervised response tokens). Left: Data Distribution within Each Category. The outer circle shows the distribution of the four education-specific capability categories and the general-purpose instruction data, and the inner circle shows the distribution of source subsets. Right: The detailed quantities of datasets.}
    \label{fig:dataset_composition}
\end{figure*}

\subsection{Preliminary Diversity Selection and Fine-grained Quality Filtering}

Before applying the more expensive quality scorer, we first reduce a small number of highly overrepresented sources whose candidate sizes substantially exceed their intended contributions to the final training mixture. 
For these sources, we use k-center greedy over BGE-M3 embeddings to select a semantically diverse subset rather than randomly subsampling the data. 
For example, RACE is reduced from 60,180 to 5,000 examples, AquilaEdu from 23,226 to 2,000, and CJEval from 17,178 to 2,000. 
Sources without substantial redundancy are retained without this preliminary compression.

We then use GPT-5.6-Terra to perform fine-grained scoring. 
The scoring rubrics refine the criteria used in Section~\ref{sec:llm-audit} into more detailed, task-specific dimensions. 
Each applicable dimension is scored on a 1--5 scale. 
For example, problem-solving data are evaluated separately for problem validity, answer correctness, reasoning correctness, completeness, relevance, and clarity.

An example is retained only if all applicable dimensions score at least 3, with a stricter threshold of 4 for task-critical dimensions such as correctness, validity, and grounding.
After this stage, the education-specific candidate pool contains 121,318 examples.

\subsection{Token-budgeted Diversity Selection}
The remaining data are high-quality but still unevenly distributed across tasks and sources. 
We therefore perform the final selection independently within the fine-grained task buckets defined in the supplementary data documentation, again using k-center greedy over BGE-M3 embeddings to maximize semantic coverage within each bucket.

We allocate each bucket a budget in terms of supervised response tokens, since response lengths differ substantially across tasks: a long tutoring dialogue can contribute an order of magnitude more training tokens than a short answer or hint. 
Within each bucket, examples are selected greedily until the corresponding supervised-token budget is reached.

This bucket-wise selection reduces the education-specific pool to \textbf{60,951 examples}, containing approximately \textbf{12.0M supervised response tokens}.

\subsection{Pedagogical Instruction Assignment and Final Assembly}
Different educational tasks require different response behaviors. 
For example, a problem-solving response should generally provide and justify the final answer, whereas a scaffolding response may deliberately withhold the answer and instead guide the student through intermediate reasoning. 
A single generic system instruction would therefore conflate distinct, and sometimes conflicting, behaviors.

Following the idea of \textit{pedagogical instruction following} in LearnLM~\cite{learnlm}, we associate each selected example with a task-specific system instruction that explicitly describes the desired response behavior. 
The instruction is determined by the example's predefined task bucket. 
We use 20 system-instruction templates covering behaviors such as reasoned problem solving, reading comprehension, diagnosis and correction, Socratic questioning, and tutoring dialogue. 
This design allows the same model to learn multiple pedagogical behaviors while making the desired behavior explicit in the input. Each example receives exactly one system instruction, while its original user and assistant content remains unchanged. 

Finally, all examples are converted to a unified message-based format, and referenced images are materialized and aligned with their corresponding image placeholders.
We perform final integrity checks on message structure, media availability, duplicate identifiers, and cross-category duplicates. 

\subsection{Final Data Statistics}

Figure~\ref{fig:dataset_composition} summarizes the final training mixture. 
The education-specific component contains 60,951 examples and approximately 12.0M supervised response tokens. 
Together with the 9,048 general-purpose examples, the complete corpus contains \textbf{69,999 examples and 15.96M supervised response tokens}.




\section{Experiments}
\label{sec:experiments}

We evaluate whether our education-oriented instruction tuning improves the capabilities required by a practical K--12 educational assistant.
Our evaluation focuses on three complementary aspects:
(1) \textit{curriculum grounding}, which measures whether the model understands knowledge points and their structural relations within a K--12 curriculum;
(2) \textit{K--12 problem solving}, which measures the model's ability to solve authentic school-level problems across subjects and modalities; and
(3) \textit{pedagogical tutoring}, which evaluates whether the model can provide effective explanations, feedback, and instructional guidance rather than merely produce correct answers.
Finally, we evaluate general-purpose benchmarks to verify that specialization toward education does not degrade the model's general capabilities.

\subsection{Experimental Setup}
\label{sec:exp_setup}

\paragraph{Models.}
We fine-tune three base models (Qwen3.5-4B-Base, Qwen3.5-9B-Base, and Qwen3.8-27B) with different scales of parameters using the education-oriented instruction corpus described in Section~\ref{sec:data}.
We denote the resulting models as \model-4B, \model-9B, and \model-27B, respectively.
Unless otherwise specified, all models are evaluated using the same prompting and decoding protocol.
We use LlamaFactory, with a learning rate of $5\times 10^{-6}$, a maximum sequence length of 32,768, and train the models for 3 epochs.
Additional training details are provided in the supplementary material.

\begin{table*}[t]
\centering
\small
\setlength{\tabcolsep}{4pt}
\caption{Curriculum-grounding results across all evaluated models. All values are percentages. Top-three values are shown in \rankOne{red}, \rankTwo{orange}, and \rankThree{yellow}, respectively.}
\label{tab:summary-curriculum}
\begin{tabular}{l c rrrr}
\toprule
\textbf{Model} & \textbf{Size} & \textbf{K12-Bench EM} & \textbf{K12-Bench F1} & \textbf{MathFish Acc.} & \textbf{EDUMATH MaC} \\
\midrule
\multicolumn{6}{c}{\cellcolor{model_type}\textbf{Open-weight Models}} \\
\midrule
\textit{Qwen3.5-4B-Base} & 4B & 42.72\% & 69.20\% & 80.33\% & 47.60\% \\
\textit{\model-4B} (ours) & 4B & 54.25\% & 71.75\% & 83.19\% & 68.40\% \\
\textit{Qwen3.5-9B-Base} & 9B & 48.52\% & 71.99\% & 79.66\% & 58.60\% \\
\textit{\model-9B} (ours) & 9B & \rankTwo{55.46\%} & \rankThree{73.68\%} & 83.70\% & 74.00\% \\
\textit{Qwen3.8-27B} & 27B & 52.11\% & 73.48\% & 83.54\% & 70.60\% \\
\textit{\model-27B} (ours) & 27B & \rankOne{63.12\%} & \rankOne{76.69\%} & \rankOne{85.89\%} & \rankTwo{86.95\%} \\
\midrule
\textit{Confucius3-Math} & 14B & 5.23\% & 19.23\% & 0.00\% & 3.00\% \\
\textit{MuduoLLM} & 14B & 48.24\% & 70.46\% & 82.85\% & 57.20\% \\
\textit{EduChat-SFT-Qwen2.5-7B} & 7B & 45.20\% & 67.58\% & 78.23\% & 28.00\% \\
\textit{EduChat-R1-Qwen3-8B} & 8B & 25.25\% & 49.67\% & 19.49\% & 47.00\% \\
\textit{EduChat-R1-Qwen3-32B} & 32B & 49.42\% & 71.21\% & 83.71\% & 61.20\% \\
\midrule
\multicolumn{6}{c}{\cellcolor{model_type}\textbf{Proprietary Models}} \\
\midrule
\textit{GPT-5.4} & -- & 43.23\% & 67.80\% & 80.65\% & 80.50\% \\
\textit{GPT-5.6-Sol} & -- & 48.63\% & 71.03\% & 83.68\% & \rankThree{84.92\%} \\
\textit{Claude-Opus-5} & -- & 48.48\% & 71.34\% & 83.52\% & 75.00\% \\
\textit{GLM-5.3} & -- & 43.38\% & 57.66\% & \rankThree{85.23\%} & 62.78\% \\
\textit{Kimi-K3} & -- & \rankThree{55.03\%} & \rankTwo{74.71\%} & \rankTwo{85.40\%} & \rankOne{90.00\%} \\
\bottomrule
\end{tabular}
\vspace{-2mm}
\end{table*}

\paragraph{Baselines.}
We compare against two groups of baselines.
First, we include representative \textbf{open-weight educational LLMs}, including Confucius3-Math, MuduoLLM, and multiple variants of EduChat.
These models cover different education-oriented training paradigms, including domain-specific pretraining, supervised fine-tuning, preference optimization, and reinforcement learning.
Second, we compare against \textbf{frontier proprietary models}, including GPT-5.4, GPT-5.6-Sol, Claude-Opus-5, GLM-5.3, and Kimi-K3.
We additionally compare every fine-tuned model with its corresponding base model to isolate the effect of our education-oriented training.
Since the five open-weight educational LLMs and GLM-5.3 do not support image inputs, they receive only the textual component for image-dependent examples.

\paragraph{Benchmarks.}
For \textit{curriculum grounding}, we evaluate on K12-Bench~\cite{k12kgraph}, MathFish~\cite{mathfish}, and EDUMATH~\cite{edumath}.
K12-Bench evaluates curriculum cognition through graph-derived tasks involving concept grounding, prerequisite relations, neighborhood relations, supporting evidence, and curriculum localization.
MathFish evaluates whether a model can determine the alignment between a mathematical problem and fine-grained curriculum standards.
EDUMATH further examines curriculum-grounded generation by requiring models to generate high-quality mathematical word problems conditioned on educational standards and student interests.
For \textit{K--12 problem solving}, we primarily report results on GAOKAO-Bench~\cite{gaokaobench}, EXAMS-V~\cite{examsv}, and MDK12-Bench~\cite{mdk12}.
These benchmarks cover authentic K--12 examination problems, including both open-ended and objective questions as well as multimodal inputs.
For \textit{pedagogical capability}, we evaluate on MathTutorBench~\cite{mathtutorbench}, TutorBench~\cite{tutorbench}, and LongTutor~\cite{longtutor}.
Together, they cover scaffolded instruction, pedagogical response quality, adaptive explanation, feedback, active-learning guidance, and long-term personalized tutoring.
We follow the official evaluation protocol and metric of each benchmark.
MathTutorBench evaluates the pedagogical quality of mathematical tutoring, focusing on scaffolding and pedagogical instruction following.
TutorBench evaluates tutoring quality across adaptive explanation, student feedback, and active-learning hints.
LongTutor evaluates long-term personalized tutoring with an emphasis on history-grounded diagnosis and adaptive teaching.
Finally, we use IFEval~\cite{zhou2023ifeval}, GPQA~\cite{rein2023gpqa}, and MMMU-Pro~\cite{yue2024mmmu_pro} as auxiliary diagnostics of how education-oriented fine-tuning affects general instruction following, scientific reasoning, and multimodal reasoning.

\subsection{Curriculum Grounding}
\label{sec:exp_curriculum}

\begin{table*}[t]
\centering
\small
\setlength{\tabcolsep}{4pt}
\caption{K--12 problem-solving results across all evaluated models. All values are percentages. Top-three values are shown in \rankOne{red}, \rankTwo{orange}, and \rankThree{yellow}, respectively.}
\label{tab:summary-problem-solving}
\begin{tabular}{l c rrr}
\toprule
\textbf{Model} & \textbf{Size} & \textbf{GAOKAO-Bench Full} & \textbf{EXAMS-V Overall} & \textbf{MDK12-Bench Full} \\
\midrule
\multicolumn{5}{c}{\cellcolor{model_type}\textbf{Open-weight Models}} \\
\midrule
\textit{Qwen3.5-4B-Base} & 4B & 88.98\% & 44.69\% & 35.43\% \\
\textit{\model-4B} (ours) & 4B & 90.46\% & 57.62\% & 46.36\% \\
\textit{Qwen3.5-9B-Base} & 9B & 92.94\% & 63.00\% & 44.50\% \\
\textit{\model-9B} (ours) & 9B & 93.66\% & 66.40\% & 50.80\% \\
\textit{Qwen3.8-27B} & 27B & 91.55\% & \rankThree{68.65\%} & 46.04\% \\
\textit{\model-27B} (ours) & 27B & \rankThree{94.87\%} & \rankTwo{69.52\%} & \rankTwo{57.76\%} \\
\midrule
\textit{Confucius3-Math} & 14B & 84.51\% & 21.95\% & 39.13\% \\
\textit{MuduoLLM} & 14B & 89.66\% & 21.95\% & 45.44\% \\
\textit{EduChat-SFT-Qwen2.5-7B} & 7B & 70.75\% & 21.95\% & 32.50\% \\
\textit{EduChat-R1-Qwen3-8B} & 8B & 76.08\% & 0.69\% & 41.29\% \\
\textit{EduChat-R1-Qwen3-32B} & 32B & 83.86\% & 21.95\% & 41.84\% \\
\midrule
\multicolumn{5}{c}{\cellcolor{model_type}\textbf{Proprietary Models}} \\
\midrule
\textit{GPT-5.4} & -- & 93.44\% & 35.66\% & 54.77\% \\
\textit{GPT-5.6-Sol} & -- & \rankTwo{95.96\%} & 24.82\% & 54.67\% \\
\textit{Claude-Opus-5} & -- & \rankOne{97.22\%} & 64.34\% & \rankThree{57.46\%} \\
\textit{GLM-5.3} & -- & 89.76\% & 22.29\% & 55.33\% \\
\textit{Kimi-K3} & -- & 94.86\% & \rankOne{87.29\%} & \rankOne{63.60\%} \\
\bottomrule
\end{tabular}
\vspace{-2mm}
\end{table*}


\paragraph{K12-Bench.}
As summarized in Table~\ref{tab:summary-curriculum}, instruction tuning substantially improves curriculum understanding across all model scales.
\model-27B increases the overall EM from 52.11\% to \textbf{63.12\%} and F1 from 73.48\% to \textbf{76.69\%}, achieving the best overall performance among the evaluated models.
The gains are consistent across all K12-Bench subtasks, showing improvements in both knowledge grounding and structural curriculum reasoning.

\paragraph{MathFish and EDUMATH.}
On MathFish, \model-27B improves overall accuracy from 83.54\% to \textbf{85.89\%}, achieving the best overall performance, with a particularly large gain on positive alignment cases.
On EDUMATH, the MaC scores of the 4B, 9B, and 27B models improve by 20.80, 15.40, and 16.35 points, respectively, with \model-27B reaching \textbf{86.95\%}, second only to Kimi-K3.
Together, these results demonstrate consistent gains in both directions of curriculum alignment: mapping problems to their underlying standards and generating problems from specified standards.

\paragraph{Overall findings.}
Across all three curriculum benchmarks, education-oriented tuning consistently improves performance at every model scale.
The 27B model consistently ranks among the strongest models, while the 4B and 9B variants also show substantial and consistent gains over their corresponding base models.
This cross-benchmark consistency indicates that the improvements in curriculum competence generalize across different task formulations and model scales.

\begin{table*}[t]
\centering
\scriptsize
\setlength{\tabcolsep}{2.5pt}
\caption{Pedagogical-tutoring results across all evaluated models. `MathTutor-S' and `MathTutor-SH' denote Scaffold and Scaffold-hard win rates; `LongTutor-E' and `LongTutor-T' denote Evidence and Teaching averages. Win rates and Evidence are percentages; Teaching is on the benchmark's original scale. Top-three values are shown in \rankOne{red}, \rankTwo{orange}, and \rankThree{yellow}, respectively.}
\label{tab:summary-tutoring}
\begin{tabular}{l c rrrrr}
\toprule
\textbf{Model} & \textbf{Size} & \textbf{MathTutor-S} & \textbf{MathTutor-SH} & \textbf{TutorBench} & \textbf{LongTutor-E} & \textbf{LongTutor-T} \\
\midrule
\multicolumn{7}{c}{\cellcolor{model_type}\textbf{Open-weight Models}} \\
\midrule
\textit{Qwen3.5-4B-Base} & 4B & 20.42\% & 18.36\% & 45.52\% & 25.67\% & 1.60 \\
\textit{\model-4B} (ours) & 4B & 75.79\% & \rankOne{84.77\%} & 46.67\% & 65.88\% & 2.29 \\
\textit{Qwen3.5-9B-Base} & 9B & 14.00\% & 13.67\% & 45.38\% & 5.81\% & 1.48 \\
\textit{\model-9B} (ours) & 9B & 75.26\% & 81.64\% & 48.16\% & 66.63\% & 2.66 \\
\textit{Qwen3.8-27B} & 27B & 57.16\% & 55.86\% & \rankThree{58.58\%} & 36.80\% & \rankTwo{2.74} \\
\textit{\model-27B} (ours) & 27B & \rankTwo{78.74\%} & \rankThree{83.59\%} & \rankTwo{59.42\%} & \rankTwo{78.20\%} & \rankOne{3.02} \\
\midrule
\textit{Confucius3-Math} & 14B & 27.68\% & 20.70\% & 34.72\% & 23.07\% & 1.36 \\
\textit{MuduoLLM} & 14B & 30.42\% & 23.44\% & 35.98\% & 24.31\% & 1.28 \\
\textit{EduChat-SFT-Qwen2.5-7B} & 7B & 18.21\% & 22.66\% & 21.48\% & 26.55\% & 1.21 \\
\textit{EduChat-R1-Qwen3-8B} & 8B & 11.58\% & 7.42\% & 23.71\% & 33.75\% & 1.14 \\
\textit{EduChat-R1-Qwen3-32B} & 32B & 22.21\% & 16.80\% & 25.92\% & 52.03\% & 1.56 \\
\midrule
\multicolumn{7}{c}{\cellcolor{model_type}\textbf{Proprietary Models}} \\
\midrule
\textit{GPT-5.4} & -- & 6.32\% & 1.96\% & 46.34\% & 75.50\% & 1.62 \\
\textit{GPT-5.6-Sol} & -- & 10.53\% & 17.65\% & 50.97\% & \rankThree{77.50\%} & 1.89 \\
\textit{Claude-Opus-5} & -- & \rankOne{87.89\%} & \rankTwo{84.31\%} & 54.01\% & \rankOne{84.17\%} & \rankThree{2.72} \\
\textit{GLM-5.3} & -- & \rankThree{77.89\%} & 76.92\% & 38.62\% & 77.42\% & 2.66 \\
\textit{Kimi-K3} & -- & 20.00\% & 15.38\% & \rankOne{63.65\%} & 74.73\% & 2.17 \\
\bottomrule
\end{tabular}
\vspace{-2mm}
\end{table*}

\subsection{K--12 Problem Solving}
\label{sec:exp_problem_solving}

Table~\ref{tab:summary-problem-solving} reports the headline results on GAOKAO-Bench, EXAMS-V, and MDK12-Bench; complete breakdowns are deferred to Appendix~\ref{app:detailed-results}.
On GAOKAO-Bench, \model-27B improves the full-score rate from 91.55\% to \textbf{94.87\%}, with gains on both objective and subjective questions.
On EXAMS-V, the improvement is particularly pronounced for \model-4B, whose overall accuracy increases from 44.69\% to 57.62\%.
On MDK12-Bench, \model-27B improves the overall score from 46.04\% to \textbf{57.76\%}, with open-question correctness rising substantially from 62.08\% to \textbf{78.77\%}.

\paragraph{Overall findings.}
Across all three K--12 problem-solving benchmarks, each tuned model outperforms its corresponding base model on the overall metric.
\model-27B consistently outperforms the evaluated open-weight educational baselines and ranks within the top three overall on all three benchmarks.
Although frontier proprietary models retain an advantage on some exam-style evaluations, \model-27B remains highly competitive despite its substantially smaller scale.

\subsection{Pedagogical Tutoring}
\label{sec:exp_tutoring}

\paragraph{MathTutorBench.}
As shown in Table~\ref{tab:summary-tutoring}, education-oriented tuning produces particularly large improvements in mathematical tutoring.
For \model-27B, Scaffold and Scaffold-hard win rates increase from 57.16\% and 55.86\% to \textbf{78.74\%} and \textbf{83.59\%}, respectively.
The effect is even more pronounced for the smaller models, with Scaffold win rate improving by 55.37 points for \model-4B and 61.26 points for \model-9B.

\paragraph{TutorBench.}
All three tuned models improve in overall performance, with \model-9B showing the largest gain among the smaller models from 45.38\% to 48.16\%.
\model-27B achieves the strongest overall score of \textbf{59.42\%}, ranking second behind Kimi-K3, while obtaining the best aggregate performance on text-only examples.

\paragraph{LongTutor.}
LongTutor shows particularly strong gains in the use of student learning history.
The Evidence average increases from 25.67\% to 65.88\% for \model-4B, from 5.81\% to 66.63\% for \model-9B, and from 36.80\% to \textbf{78.20\%} for \model-27B.
While Claude-Opus-5 achieves a higher Evidence score, \model-27B obtains the best Teaching average of \textbf{3.02}.
\model also performs particularly well on knowledge-state diagnosis, with \model-27B and \model-9B ranking first and second in diagnosis accuracy at \textbf{54.04\%} and \textbf{53.55\%}, respectively.
However, the relatively low accuracies indicate that knowledge-state diagnosis remains comparatively challenging.

\paragraph{Overall findings.}
Across all three tutoring benchmarks, education-oriented tuning improves pedagogical performance at every model scale.
Taken together, these results demonstrate that the benefit of our corpus extends beyond answer correctness to broader tutoring capabilities.
The largest gains appear in scaffolded instruction and the use of longitudinal student evidence.

\subsection{General Capability Retention}
\label{sec:exp_general}

In addition to the above education domain benchmarks, we verify that specializing the models for K--12 education does not come at the expense of their general capabilities.
We evaluate the base and tuned models on IFEval, GPQA, and MMMU-Pro, covering instruction following, challenging knowledge-intensive reasoning, and multimodal general understanding. 
Complete results are reported in Appendix~\ref{app:general-capability}.
Across the three scales, tuning generally improves IFEval metrics and improves both GPQA Main and Diamond accuracy. 
MMMU-Pro overall accuracy increases from 50.46\% to 52.60\% for 4B, from 58.38\% to 60.75\% for 9B, and from 64.97\% to 67.98\% for 27B. 
These results indicate that the educational specialization gains do not require a broad loss of general capability, although changes vary by MMMU-Pro discipline.


\section{Conclusion}

In this work, we present \model, a family of open foundation models for K--12 learning and teaching, trained with a capability-oriented instruction-tuning corpus.
Rather than organizing training data solely by source or subject, we structure supervision around four complementary educational capabilities: \textit{subject competence}, \textit{curriculum grounding}, \textit{diagnostic reasoning}, and \textit{pedagogical action and scaffolding}.
We construct a compact, high-quality corpus through a multi-stage pipeline that combines deterministic cleaning, semantic auditing and revision, task-specific quality assessment, and diversity-aware sampling.
Experiments across 4B, 9B, and 27B models show consistent improvements across model scales.
In particular, the OmniEdu family achieves the strongest open-weight results across the curriculum-grounding and pedagogical-tutoring benchmarks, while \model-27B remains competitive with frontier proprietary models on K--12 problem solving and substantially outperforms existing open-weight educational models.
We additionally use general instruction-following, reasoning, and multimodal evaluations to quantify the broader effects of education-oriented specialization.
Overall, our results suggest that carefully curated and capability-balanced supervision provides an effective path toward educational foundation models that go beyond problem solving, enabling stronger understanding of curriculum structure and more effective pedagogical interaction.


\clearpage
\bibliographystyle{plainnat}
\bibliography{main}
\beginappendix

\setcounter{topnumber}{4}
\setcounter{bottomnumber}{2}
\setcounter{totalnumber}{6}
\renewcommand{\topfraction}{0.92}
\renewcommand{\bottomfraction}{0.85}
\renewcommand{\textfraction}{0.08}
\renewcommand{\floatpagefraction}{0.78}
\setlength{\textfloatsep}{12pt plus 2pt minus 2pt}
\setlength{\floatsep}{10pt plus 2pt minus 2pt}
\setlength{\intextsep}{10pt plus 2pt minus 2pt}

\section{Detailed Evaluation Results}
\label{app:detailed-results}

This appendix reports the complete model-by-model results underlying the compact summary tables in the main text. The original table structure is retained to make every metric and baseline directly auditable.

\subsection{Curriculum Grounding}
Tables~\ref{tab:k12-bench}, \ref{tab:mathfish}, and \ref{tab:edumath} provide detailed results on the three curriculum-grounding benchmarks.
K12-Bench reports performance across different types of curriculum relations, MathFish separates positive and negative curriculum-alignment cases, and EDUMATH reports complementary metrics for curriculum-grounded problem generation.
Overall, the detailed results show that the improvements after education-oriented tuning extend across multiple forms of curriculum understanding rather than being driven by a single evaluation setting.

%

\begin{table*}[t]
\centering
\caption{Detailed K12-Bench results across curriculum-grounding subtasks. Top-three values are shown in \rankOne{red}, \rankTwo{orange}, and \rankThree{yellow}, respectively.}
\label{tab:k12-bench}
\resizebox{\linewidth}{!}{
\begin{tabular}{l c cc cc cc cc cc cc}
\toprule
\multirow{2}{*}{\textbf{Model}} & \multirow{2}{*}{\textbf{Size}} &
\multicolumn{2}{c}{\textsc{Ground}} &
\multicolumn{2}{c}{\textsc{Prereq}} &
\multicolumn{2}{c}{\textsc{Neighbor}} &
\multicolumn{2}{c}{\textsc{Evidence}} &
\multicolumn{2}{c}{\textsc{Locate}} &
\multicolumn{2}{c}{\textbf{Overall}} \\
\cmidrule(lr){3-4} \cmidrule(lr){5-6} \cmidrule(lr){7-8} \cmidrule(lr){9-10} \cmidrule(lr){11-12} \cmidrule(lr){13-14}
& & \textbf{EM} & \textbf{F1} & \textbf{EM} & \textbf{F1} & \textbf{EM} & \textbf{F1} &
\textbf{EM} & \textbf{F1} & \textbf{EM} & \textbf{F1} & \textbf{EM} & \textbf{F1} \\
\midrule
\multicolumn{14}{c}{\cellcolor{model_type}\textbf{Open-weight Models}} \\
\midrule
\textit{Qwen3.5-4B-Base} & 4B & 47.16\% & 78.65\% & 41.53\% & 73.44\% & 66.83\% & 71.52\% & 28.48\% & 69.61\% & 15.53\% & 59.73\% & 42.72\% & 69.20\% \\
\textit{\model-4B} (ours) & 4B & 59.74\% & 82.59\% & 60.55\% & 79.26\% & 70.51\% & 71.13\% & 37.95\% & 72.71\% & 35.59\% & 63.86\% & 54.25\% & 71.75\% \\
\textit{Qwen3.5-9B-Base} & 9B & 55.93\% & 81.86\% & 50.03\% & 77.65\% & 70.44\% & 73.94\% & 31.51\% & 70.91\% & 24.33\% & 63.21\% & 48.52\% & 71.99\% \\
\textit{\model-9B} (ours) & 9B & 60.46\% & 83.34\% & 62.92\% & 82.26\% & 72.68\% & 74.13\% & 38.00\% & 72.94\% & 35.50\% & 64.98\% & \rankTwo{55.46\%} & \rankThree{73.68\%} \\
\textit{Qwen3.8-27B} & 27B & 54.35\% & 81.23\% & 60.32\% & 83.56\% & 74.13\% & 74.90\% & 28.22\% & 70.37\% & 29.85\% & 64.59\% & 52.11\% & 73.48\% \\
\textit{\model-27B} (ours) & 27B & 70.80\% & 87.65\% & 73.67\% & 87.54\% & 75.85\% & 76.37\% & 46.25\% & 76.57\% & 48.77\% & 67.82\% & \rankOne{63.12\%} & \rankOne{76.69\%} \\
\textit{Confucius3-Math} & 14B & 5.68\% & 22.46\% & 2.99\% & 11.79\% & 7.12\% & 24.99\% & 1.47\% & 5.11\% & 6.74\% & 25.86\% & 5.23\% & 19.23\% \\
\textit{MuduoLLM} & 14B & 47.95\% & 77.66\% & 54.29\% & 78.53\% & 70.29\% & 72.01\% & 29.00\% & 69.04\% & 24.37\% & 61.54\% & 48.24\% & 70.46\% \\
\textit{EduChat-SFT-Qwen2.5-7B} & 7B & 50.40\% & 78.34\% & 52.78\% & 78.19\% & 63.89\% & 66.90\% & 33.50\% & 69.59\% & 18.13\% & 56.73\% & 45.20\% & 67.58\% \\
\textit{EduChat-R1-Qwen3-8B} & 8B & 30.55\% & 59.12\% & 23.83\% & 58.45\% & 40.39\% & 55.09\% & 15.14\% & 52.56\% & 8.93\% & 30.03\% & 25.25\% & 49.67\% \\
\textit{EduChat-R1-Qwen3-32B} & 32B & 56.29\% & 81.58\% & 58.00\% & 81.44\% & 71.39\% & 72.39\% & 29.05\% & 69.12\% & 22.96\% & 61.08\% & 49.42\% & 71.21\% \\
\midrule
\multicolumn{14}{c}{\cellcolor{model_type}\textbf{Proprietary Models}} \\
\midrule
\textit{GPT-5.4} & -- & 43.17\% & 74.42\% & 48.31\% & 77.71\% & 70.14\% & 71.16\% & 11.35\% & 59.59\% & 22.44\% & 60.31\% & 43.23\% & 67.80\% \\
\textit{GPT-5.6-Sol} & -- & 43.53\% & 74.83\% & 50.00\% & 74.81\% & 78.78\% & 79.72\% & 11.24\% & 58.59\% & 31.02\% & 63.54\% & 48.63\% & 71.03\% \\
\textit{Claude-Opus-5} & -- & 43.17\% & 75.69\% & 53.25\% & 81.03\% & 75.65\% & 76.63\% & 14.68\% & 61.17\% & 30.38\% & 63.09\% & 48.48\% & 71.34\% \\
\textit{GLM-5.3} & -- & 43.53\% & 62.29\% & 55.32\% & 72.02\% & 64.81\% & 65.31\% & 12.50\% & 44.14\% & 25.58\% & 44.94\% & 43.38\% & 57.66\% \\
\textit{Kimi-K3} & -- & 57.19\% & 80.79\% & 63.25\% & 82.21\% & 81.97\% & 81.99\% & 15.25\% & 61.76\% & 37.77\% & 66.63\% & \rankThree{55.03\%} & \rankTwo{74.71\%} \\
\bottomrule
\end{tabular}}
\vspace{-2mm}
\end{table*}
\begin{table*}[t]
\centering
\caption{Detailed MathFish results across positive and negative curriculum-alignment cases. Top-three values are shown in \rankOne{red}, \rankTwo{orange}, and \rankThree{yellow}, respectively.}
\label{tab:mathfish}
\resizebox{\linewidth}{!}{
\begin{tabular}{l c c c cc cc c}
\toprule
\multirow{2}{*}{\textbf{Model}} &
\multirow{2}{*}{\textbf{Size}} &
\multirow{2}{*}{\textsc{Pos: Should Yes}} &
\multirow{2}{*}{\textsc{Neg: ATC Neighbor}} &
\multicolumn{2}{c}{\textsc{Neg: Same-domain}} &
\multicolumn{2}{c}{\textsc{Neg: Diff-domain}} &
\multirow{2}{*}{\textbf{Overall Acc}} \\
\cmidrule(lr){5-6}
\cmidrule(lr){7-8}
& & & &
\textbf{Same-grade} & \textbf{Diff-grade} &
\textbf{Same-grade} & \textbf{Diff-grade} & \\
\midrule
\multicolumn{9}{c}{\cellcolor{model_type}\textbf{Open-weight Models}} \\
\midrule
\textit{Qwen3.5-4B-Base} & 4B & 51.20\% & 77.70\% & 86.40\% & 90.70\% & 97.10\% & 98.60\% & 80.33\% \\
\textit{\model-4B} (ours) & 4B & 63.74\% & 73.22\% & 87.91\% & 89.50\% & 98.25\% & 99.64\% & 83.19\% \\
\textit{Qwen3.5-9B-Base} & 9B & 33.40\% & 88.10\% & 93.40\% & 95.90\% & 99.10\% & 99.50\% & 79.66\% \\
\textit{\model-9B} (ours) & 9B & 62.00\% & 76.50\% & 89.80\% & 91.10\% & 98.10\% & 99.50\% & 83.70\% \\
\textit{Qwen3.8-27B} & 27B & 45.95\% & 88.31\% & 95.86\% & 97.01\% & 99.64\% & 99.95\% & 83.54\% \\
\textit{\model-27B} (ours) & 27B & 67.60\% & 79.07\% & 89.50\% & 92.67\% & 98.97\% & 99.74\% & \rankOne{85.89\%} \\
\textit{Confucius3-Math} & 14B & 0.00\% & 0.20\% & 0.10\% & 0.10\% & 0.10\% & 0.10\% & 0.00\% \\
\textit{MuduoLLM} & 14B & 66.10\% & 71.00\% & 84.40\% & 89.70\% & 98.00\% & 99.20\% & 82.85\% \\
\textit{EduChat-SFT-Qwen2.5-7B} & 7B & 42.00\% & 78.20\% & 87.00\% & 90.40\% & 97.50\% & 98.70\% & 78.23\% \\
\textit{EduChat-R1-Qwen3-8B} & 8B & 12.30\% & 22.30\% & 32.00\% & 38.20\% & 49.70\% & 55.30\% & 19.49\% \\
\textit{EduChat-R1-Qwen3-32B} & 32B & 74.70\% & 66.50\% & 82.00\% & 88.80\% & 97.50\% & 99.00\% & 83.71\% \\
\midrule
\multicolumn{9}{c}{\cellcolor{model_type}\textbf{Proprietary Models}} \\
\midrule
\textit{GPT-5.4} & -- & 73.49\% & 61.76\% & 76.90\% & 81.68\% & 96.40\% & 98.20\% & 80.65\% \\
\textit{GPT-5.6-Sol} & -- & 65.27\% & 75.97\% & 86.88\% & 90.58\% & 96.92\% & 98.71\% & 83.68\% \\
\textit{Claude-Opus-5} & -- & 52.71\% & 83.98\% & 93.70\% & 93.46\% & 98.71\% & 99.23\% & 83.52\% \\
\textit{GLM-5.3} & -- & 66.67\% & 77.78\% & 89.50\% & 91.62\% & 98.97\% & 99.50\% & \rankThree{85.23\%} \\
\textit{Kimi-K3} & -- & 68.60\% & 75.50\% & 89.40\% & 92.60\% & 98.30\% & 99.50\% & \rankTwo{85.40\%} \\
\bottomrule
\end{tabular}}
\vspace{-2mm}
\end{table*}
\begin{table*}[t]
\centering
\caption{Detailed EDUMATH results on curriculum-grounded problem generation. Top-three values are shown in \rankOne{red}, \rankTwo{orange}, and \rankThree{yellow}, respectively.}
\label{tab:edumath}
\resizebox{0.72\linewidth}{!}{
\begin{tabular}{l c c c c}
\toprule
\textbf{Model} & \textbf{Size} & \textbf{Format Pass} & \textbf{Classifier GOOD} & \textbf{MaC} \\
\midrule
\multicolumn{5}{c}{\cellcolor{model_type}\textbf{Open-weight Models}} \\
\midrule
\textit{Qwen3.5-4B-Base} & 4B & 95.90\% & 86.90\% & 47.60\% \\
\textit{\model-4B} (ours) & 4B & 99.20\% & 90.20\% & 68.40\% \\
\textit{Qwen3.5-9B-Base} & 9B & 98.10\% & 89.20\% & 58.60\% \\
\textit{\model-9B} (ours) & 9B & 99.80\% & 91.70\% & 74.00\% \\
\textit{Qwen3.8-27B} & 27B & 100.00\% & 91.00\% & 70.60\% \\
\textit{\model-27B} (ours) & 27B & 98.80\% & 92.60\% & \rankTwo{86.95\%} \\
\textit{Confucius3-Math} & 14B & 11.70\% & 91.90\% & 3.00\% \\
\textit{MuduoLLM} & 14B & 92.70\% & 90.60\% & 57.20\% \\
\textit{EduChat-SFT-Qwen2.5-7B} & 7B & 15.60\% & 96.10\% & 28.00\% \\
\textit{EduChat-R1-Qwen3-8B} & 8B & 36.60\% & 93.40\% & 47.00\% \\
\textit{EduChat-R1-Qwen3-32B} & 32B & 67.60\% & 91.80\% & 61.20\% \\
\midrule
\multicolumn{5}{c}{\cellcolor{model_type}\textbf{Proprietary Models}} \\
\midrule
\textit{GPT-5.4} & -- & 100.00\% & 94.50\% & 80.50\% \\
\textit{GPT-5.6-Sol} & -- & 99.50\% & 89.45\% & \rankThree{84.92\%} \\
\textit{Claude-Opus-5} & -- & 100.00\% & 75.50\% & 75.00\% \\
\textit{GLM-5.3} & -- & 96.77\% & 87.78\% & 62.78\% \\
\textit{Kimi-K3} & -- & 100.00\% & 84.00\% & \rankOne{90.00\%} \\
\bottomrule
\end{tabular}}
\vspace{-2mm}
\end{table*}

\subsection{K--12 Problem Solving}
Tables~\ref{tab:gaokao}, \ref{tab:exams-v}, and \ref{tab:mdk12} provide the complete breakdowns for the three K--12 problem-solving benchmarks.
GAOKAO-Bench reports results by question type and school subject, EXAMS-V further distinguishes modalities, subject groups, and languages, and MDK12-Bench separates objective and open-ended questions.
The breakdowns show that the gains from education-oriented tuning generally extend across different subjects, modalities, and problem formats, although the magnitude of improvement varies across individual subsets.

\begin{table*}[t]
\centering
\caption{Detailed GAOKAO-Bench results by question type and school subject. Top-three values are shown in \rankOne{red}, \rankTwo{orange}, and \rankThree{yellow}, respectively.}
\label{tab:gaokao}
\resizebox{\linewidth}{!}{
\begin{tabular}{l c ccc ccccccccc}
\toprule
\textbf{Model} & \textbf{Size} & \textbf{Obj Rate} & \textbf{Subj Rate} & \textbf{Full Rate} &
\textbf{Chinese} & \textbf{English} & \textbf{Math} & \textbf{Physics} & \textbf{Chemistry} &
\textbf{Biology} & \textbf{Politics} & \textbf{History} & \textbf{Geography} \\
\midrule
\multicolumn{14}{c}{\cellcolor{model_type}\textbf{Open-weight Models}} \\
\midrule
\textit{Qwen3.5-4B-Base} & 4B & 85.00\% & 92.07\% & 88.98\% & 90.89\% & 85.32\% & 84.96\% & 85.37\% & 87.44\% & 91.51\% & 95.29\% & 93.12\% & 85.15\% \\
\textit{\model-4B} (ours) & 4B & 87.10\% & 93.19\% & 90.46\% & 59.30\% & 86.40\% & 94.70\% & 86.70\% & 83.10\% & 91.30\% & 92.50\% & 81.50\% & 82.10\% \\
\textit{Qwen3.5-9B-Base} & 9B & 90.70\% & 94.79\% & 92.94\% & 77.20\% & 88.50\% & 92.10\% & 92.20\% & 89.50\% & 97.30\% & 95.30\% & 89.90\% & 84.20\% \\
\textit{\model-9B} (ours) & 9B & 91.50\% & 95.44\% & 93.66\% & 93.37\% & 91.63\% & 90.59\% & 92.37\% & 91.56\% & 95.15\% & 96.39\% & 96.69\% & 92.73\% \\
\textit{Qwen3.8-27B} & 27B & 84.00\% & 97.73\% & 91.55\% & 55.10\% & 63.20\% & 92.80\% & 86.70\% & 83.90\% & 96.70\% & 96.60\% & 95.80\% & 82.10\% \\
\textit{\model-27B} (ours) & 27B & 91.03\% & \rankThree{98.02\%} & \rankThree{94.87\%} & 94.26\% & 85.75\% & 96.09\% & 93.37\% & 93.14\% & 97.79\% & 98.48\% & 97.64\% & 93.03\% \\
\textit{Confucius3-Math} & 14B & 70.50\% & 96.00\% & 84.51\% & 38.90\% & 57.00\% & 80.10\% & 21.10\% & 57.30\% & 88.00\% & 87.80\% & 86.80\% & 62.10\% \\
\textit{MuduoLLM} & 14B & 87.30\% & 91.60\% & 89.66\% & 76.00\% & 89.50\% & 90.30\% & 70.30\% & 70.20\% & 90.70\% & 95.30\% & 88.20\% & 85.30\% \\
\textit{EduChat-SFT-Qwen2.5-7B} & 7B & 66.60\% & 74.13\% & 70.75\% & 58.70\% & 79.60\% & 54.60\% & 37.50\% & 46.00\% & 73.30\% & 85.30\% & 70.70\% & 48.40\% \\
\textit{EduChat-R1-Qwen3-8B} & 8B & 73.70\% & 78.00\% & 76.08\% & 61.70\% & 83.30\% & 67.80\% & 50.00\% & 55.60\% & 82.00\% & 83.10\% & 72.50\% & 80.00\% \\
\textit{EduChat-R1-Qwen3-32B} & 32B & 87.20\% & 81.15\% & 83.86\% & 77.20\% & 93.70\% & 80.30\% & 72.70\% & 73.40\% & 93.30\% & 92.50\% & 91.30\% & 97.90\% \\
\midrule
\multicolumn{14}{c}{\cellcolor{model_type}\textbf{Proprietary Models}} \\
\midrule
\textit{GPT-5.4} & -- & 88.09\% & 97.83\% & 93.44\% & 90.02\% & 93.06\% & 97.27\% & 86.11\% & 76.67\% & 95.01\% & 94.68\% & 93.90\% & 96.53\% \\
\textit{GPT-5.6-Sol} & -- & \rankThree{95.67\%} & \rankTwo{98.89\%} & \rankTwo{95.96\%} & 99.56\% & 90.00\% & 99.15\% & 95.45\% & 100.00\% & 100.00\% & 100.00\% & 100.00\% & 100.00\% \\
\textit{Claude-Opus-5} & -- & \rankOne{96.96\%} & \rankOne{99.07\%} & \rankOne{97.22\%} & 98.54\% & 95.56\% & 98.86\% & 100.00\% & 100.00\% & 100.00\% & 99.63\% & 100.00\% & 100.00\% \\
\textit{GLM-5.3} & -- & 93.35\% & 96.29\% & 89.76\% & 96.66\% & 99.33\% & 91.14\% & 82.35\% & 98.33\% & 98.93\% & 99.63\% & 98.94\% & 94.00\% \\
\textit{Kimi-K3} & -- & \rankTwo{96.16\%} & 93.81\% & 94.86\% & 95.82\% & 92.67\% & 89.10\% & 89.39\% & 90.00\% & 98.22\% & 92.22\% & 98.42\% & 100.00\% \\
\bottomrule
\end{tabular}}
\vspace{-2mm}
\end{table*}
\begin{table*}[t]
\centering
\caption{Detailed EXAMS-V results across modalities, subject groups, and languages. Top-three values in Overall, Text, and Image+Text are shown in \rankOne{red}, \rankTwo{orange}, and \rankThree{yellow}, respectively.}
\label{tab:exams-v}
\resizebox{\linewidth}{!}{
\begin{tabular}{l c cccccccc}
\toprule
\textbf{Model} & \textbf{Size} & \textbf{Overall} & \textbf{Text} & \textbf{Image+Text} &
\textbf{Natural Science} & \textbf{Social Sciences} & \textbf{Other} & \textbf{Chinese} & \textbf{English} \\
\midrule
\multicolumn{10}{c}{\cellcolor{model_type}\textbf{Open-weight Models}} \\
\midrule
\textit{Qwen3.5-4B-Base} & 4B & 44.69\% & 48.57\% & 33.33\% & 41.31\% & 53.03\% & 36.68\% & 21.67\% & 23.05\% \\
\textit{\model-4B} (ours) & 4B & 57.62\% & 59.00\% & 53.56\% & 56.16\% & 64.46\% & 44.71\% & 50.17\% & 41.21\% \\
\textit{Qwen3.5-9B-Base} & 9B & 63.00\% & 64.68\% & 58.07\% & 60.65\% & 71.58\% & 49.27\% & 58.00\% & 40.35\% \\
\textit{\model-9B} (ours) & 9B & 66.40\% & 68.04\% & 61.59\% & 64.73\% & 73.58\% & 53.47\% & 59.33\% & 48.99\% \\
\textit{Qwen3.8-27B} & 27B &
\rankThree{68.65\%} &
\rankThree{69.02\%} &
\rankThree{67.57\%} &
69.22\% & 72.83\% & 53.65\% & 65.50\% & 42.94\% \\
\textit{\model-27B} (ours) & 27B &
\rankTwo{69.52\%} &
\rankTwo{69.97\%} &
\rankTwo{68.22\%} &
70.62\% & 73.52\% & 52.55\% & 67.17\% & 45.24\% \\
\textit{Confucius3-Math} & 14B & 21.95\% & 22.26\% & 21.05\% & 21.37\% & 23.24\% & 20.99\% & 21.33\% & 25.94\% \\
\textit{MuduoLLM} & 14B & 21.95\% & 22.26\% & 21.05\% & 21.37\% & 23.24\% & 20.99\% & 21.33\% & 25.94\% \\
\textit{EduChat-SFT-Qwen2.5-7B} & 7B & 21.95\% & 22.26\% & 21.05\% & 21.37\% & 23.24\% & 20.99\% & 21.33\% & 25.94\% \\
\textit{EduChat-R1-Qwen3-8B} & 8B & 0.69\% & 0.92\% & 0.00\% & 0.72\% & 0.37\% & 1.46\% & 0.00\% & 0.00\% \\
\textit{EduChat-R1-Qwen3-32B} & 32B & 21.95\% & 22.26\% & 21.05\% & 21.37\% & 23.24\% & 20.99\% & 21.33\% & 25.94\% \\
\midrule
\multicolumn{10}{c}{\cellcolor{model_type}\textbf{Proprietary Models}} \\
\midrule
\textit{GPT-5.4} & -- & 35.66\% & 35.89\% & 34.98\% & 36.11\% & 37.93\% & 27.03\% & 44.17\% & 23.19\% \\
\textit{GPT-5.6-Sol} & -- & 24.82\% & 24.86\% & 24.69\% & 24.76\% & 26.02\% & 21.62\% & 31.67\% & 18.84\% \\
\textit{Claude-Opus-5} & -- & 64.34\% & 64.94\% & 62.55\% & 68.05\% & 63.01\% & 50.45\% & 64.17\% & 59.42\% \\
\textit{GLM-5.3} & -- & 22.29\% & 22.83\% & 20.90\% & 23.88\% & 21.52\% & 16.67\% & 21.74\% & 27.27\% \\
\textit{Kimi-K3} & -- &
\rankOne{87.29\%} &
\rankOne{86.99\%} &
\rankOne{88.06\%} &
90.30\% & 85.44\% & 77.78\% & 92.75\% & 81.82\% \\
\bottomrule
\end{tabular}}
\vspace{-2mm}
\end{table*}
\begin{table*}[t]
\centering
\caption{Detailed MDK12-Bench results across objective and open-ended question types. Top-three values in Objective Avg, Open Correctness, and Full Overall are shown in \rankOne{red}, \rankTwo{orange}, and \rankThree{yellow}, respectively.}
\label{tab:mdk12}
\resizebox{\linewidth}{!}{
\begin{tabular}{l c cccccc}
\toprule
\textbf{Model} & \textbf{Size} & \textbf{Single-choice} & \textbf{Multiple-choice} &
\textbf{True/False} & \textbf{Objective Avg} & \textbf{Open Correctness} & \textbf{Full Overall} \\
\midrule
\multicolumn{8}{c}{\cellcolor{model_type}\textbf{Open-weight Models}} \\
\midrule
\textit{Qwen3.5-4B-Base} & 4B & 30.99\% & 9.79\% & 28.43\% & 23.17\% & 49.54\% & 35.43\% \\
\textit{\model-4B} (ours) & 4B & 39.90\% & 7.60\% & 38.46\% & 28.66\% & 66.74\% & 46.36\% \\
\textit{Qwen3.5-9B-Base} & 9B & 49.92\% & 10.11\% & 18.52\% & 27.94\% & 63.56\% & 44.50\% \\
\textit{\model-9B} (ours) & 9B & 49.60\% & 15.98\% & 38.86\% & 35.37\% & 68.55\% & 50.80\% \\
\textit{Qwen3.8-27B} & 27B & 56.34\% & 13.42\% & 17.32\% & 32.10\% & 62.08\% & 46.04\% \\
\textit{\model-27B} (ours) & 27B &
53.97\% & 25.02\% & 31.74\% &
38.15\% &
\rankTwo{78.77\%} &
\rankTwo{57.76\%} \\
\textit{Confucius3-Math} & 14B & 34.16\% & 4.19\% & 14.53\% & 18.72\% & 62.63\% & 39.13\% \\
\textit{MuduoLLM} & 14B & 55.57\% & 11.60\% & 20.23\% & 31.12\% & 61.92\% & 45.44\% \\
\textit{EduChat-SFT-Qwen2.5-7B} & 7B & 37.45\% & 7.97\% & 10.77\% & 20.23\% & 46.62\% & 32.50\% \\
\textit{EduChat-R1-Qwen3-8B} & 8B & 52.65\% & 4.94\% & 25.36\% & 29.15\% & 55.27\% & 41.29\% \\
\textit{EduChat-R1-Qwen3-32B} & 32B & 52.53\% & 12.26\% & 7.58\% & 26.68\% & 59.30\% & 41.84\% \\
\midrule
\multicolumn{8}{c}{\cellcolor{model_type}\textbf{Proprietary Models}} \\
\midrule
\textit{GPT-5.4} & -- &
44.44\% & 35.20\% & 32.47\% &
38.05\% & 73.80\% & 54.77\% \\
\textit{GPT-5.6-Sol} & -- &
51.11\% & 41.26\% & 29.02\% &
\rankTwo{41.75\%} &
69.82\% &
54.67\% \\
\textit{Claude-Opus-5} & -- &
61.82\% & 23.08\% & 22.13\% &
37.89\% &
\rankThree{78.43\%} &
\rankThree{57.46\%} \\
\textit{GLM-5.3} & -- &
57.08\% & 31.03\% & 30.86\% &
\rankThree{41.10\%} &
70.57\% &
55.33\% \\
\textit{Kimi-K3} & -- &
66.67\% & 29.06\% & 32.57\% &
\rankOne{44.66\%} &
\rankOne{83.89\%} &
\rankOne{63.60\%} \\
\bottomrule
\end{tabular}}
\vspace{-2mm}
\end{table*}

\subsection{Pedagogical Tutoring}
Tables~\ref{tab:mathtutor}, \ref{tab:tutorbench}, and \ref{tab:longtutor} provide fine-grained results for the three pedagogical-tutoring benchmarks.
MathTutorBench evaluates scaffolded and pedagogy-oriented tutoring under standard and harder settings, TutorBench covers multiple tutoring use cases in both text and multimodal settings, and LongTutor evaluates evidence use, knowledge-state diagnosis, and teaching quality over longer interaction histories.
Across these complementary settings, education-oriented tuning consistently strengthens the models' ability to provide pedagogically appropriate and context-sensitive tutoring responses.

\begin{table*}[t]
\centering
\caption{Detailed MathTutorBench results across scaffolding and pedagogy settings. Top-three values are shown in \rankOne{red}, \rankTwo{orange}, and \rankThree{yellow}, respectively.}
\label{tab:mathtutor}
\resizebox{0.82\linewidth}{!}{
\begin{tabular}{l c cccc}
\toprule
\textbf{Model} & \textbf{Size} & \textbf{Scaffold WR} & \textbf{Scaffold-hard WR} &
\textbf{Pedagogy WR} & \textbf{Pedagogy-hard WR} \\
\midrule
\multicolumn{6}{c}{\cellcolor{model_type}\textbf{Open-weight Models}} \\
\midrule
\textit{Qwen3.5-4B-Base} & 4B & 20.42\% & 18.36\% & 37.05\% & 50.78\% \\
\textit{\model-4B} (ours) & 4B &
75.79\% &
\rankOne{84.77\%} &
76.53\% &
\rankTwo{86.33\%} \\
\textit{Qwen3.5-9B-Base} & 9B & 14.00\% & 13.67\% & 55.47\% & 56.64\% \\
\textit{\model-9B} (ours) & 9B & 75.26\% & 81.64\% & 78.00\% & 80.47\% \\
\textit{Qwen3.8-27B} & 27B & 57.16\% & 55.86\% & 74.74\% & 73.83\% \\
\textit{\model-27B} (ours) & 27B &
\rankTwo{78.74\%} &
\rankThree{83.59\%} &
\rankTwo{79.79\%} &
82.81\% \\
\textit{Confucius3-Math} & 14B & 27.68\% & 20.70\% & 28.74\% & 25.39\% \\
\textit{MuduoLLM} & 14B & 30.42\% & 23.44\% & 50.95\% & 37.50\% \\
\textit{EduChat-SFT-Qwen2.5-7B} & 7B & 18.21\% & 22.66\% & 20.21\% & 28.52\% \\
\textit{EduChat-R1-Qwen3-8B} & 8B & 11.58\% & 7.42\% & 33.58\% & 20.70\% \\
\textit{EduChat-R1-Qwen3-32B} & 32B & 22.21\% & 16.80\% & 51.68\% & 33.20\% \\
\midrule
\multicolumn{6}{c}{\cellcolor{model_type}\textbf{Proprietary Models}} \\
\midrule
\textit{GPT-5.4} & -- & 6.32\% & 1.96\% & 33.16\% & 41.18\% \\
\textit{GPT-5.6-Sol} & -- & 10.53\% & 17.65\% & 60.00\% & 56.86\% \\
\textit{Claude-Opus-5} & -- &
\rankOne{87.89\%} &
\rankTwo{84.31\%} &
\rankThree{78.95\%} &
\rankOne{88.24\%} \\
\textit{GLM-5.3} & -- &
\rankThree{77.89\%} &
76.92\% &
\rankOne{83.16\%} &
\rankThree{84.62\%} \\
\textit{Kimi-K3} & -- & 20.00\% & 15.38\% & 69.47\% & 69.23\% \\
\bottomrule
\end{tabular}}
\vspace{-2mm}
\end{table*}
\begin{table*}[t]
\centering
\caption{Detailed TutorBench results across tutoring use cases and input modalities. Top-three values are shown in \rankOne{red}, \rankTwo{orange}, and \rankThree{yellow}, respectively.}
\label{tab:tutorbench}
\resizebox{\linewidth}{!}{
\begin{tabular}{l c cc cc cc ccc}
\toprule
\multirow{2}{*}{\textbf{Model}} &
\multirow{2}{*}{\textbf{Size}} &
\multicolumn{2}{c}{\textsc{UC1}} &
\multicolumn{2}{c}{\textsc{UC2}} &
\multicolumn{2}{c}{\textsc{UC3}} &
\multirow{2}{*}{\textbf{Text}} &
\multirow{2}{*}{\textbf{Multimodal}} &
\multirow{2}{*}{\textbf{Overall}} \\
\cmidrule(lr){3-4}
\cmidrule(lr){5-6}
\cmidrule(lr){7-8}
& & \textbf{Text} & \textbf{Multimodal} &
\textbf{Text} & \textbf{Multimodal} &
\textbf{Text} & \textbf{Multimodal} & & & \\
\midrule
\multicolumn{11}{c}{\cellcolor{model_type}\textbf{Open-weight Models}} \\
\midrule
\textit{Qwen3.5-4B-Base} & 4B & 55.42\% & 52.75\% & 62.50\% & 38.10\% & 48.49\% & 31.01\% & 55.72\% & 37.91\% & 45.52\% \\
\textit{\model-4B} (ours) & 4B & 49.78\% & 53.66\% & 61.07\% & 36.69\% & 54.22\% & 40.47\% & 53.90\% & 41.05\% & 46.67\% \\
\textit{Qwen3.5-9B-Base} & 9B & 56.26\% & 56.89\% & 53.43\% & 38.35\% & 40.47\% & 35.05\% & 51.88\% & 40.46\% & 45.38\% \\
\textit{\model-9B} (ours) & 9B & 54.49\% & 54.88\% & 54.09\% & 41.31\% & 56.71\% & 40.24\% & 54.92\% & 43.06\% & 48.16\% \\
\textit{Qwen3.8-27B} & 27B & 69.68\% & 70.02\% & 70.68\% & 54.00\% & 54.72\% & 43.14\% & \rankTwo{66.42\%} & \rankThree{52.64\%} & \rankThree{58.58\%} \\
\textit{\model-27B} (ours) & 27B & 65.67\% & 65.49\% & 74.31\% & 54.74\% & 61.38\% & 47.87\% & \rankOne{67.00\%} & \rankTwo{53.79\%} & \rankTwo{59.42\%} \\
\textit{Confucius3-Math} & 14B & 47.64\% & 49.67\% & 50.65\% & 5.99\% & 53.48\% & 30.10\% & 49.80\% & 23.34\% & 34.72\% \\
\textit{MuduoLLM} & 14B & 50.56\% & 51.37\% & 64.25\% & 9.88\% & 51.52\% & 23.16\% & 54.32\% & 22.48\% & 35.98\% \\
\textit{EduChat-SFT-Qwen2.5-7B} & 7B & 28.62\% & 27.04\% & 28.47\% & 5.44\% & 41.95\% & 17.95\% & 31.56\% & 14.19\% & 21.48\% \\
\textit{EduChat-R1-Qwen3-8B} & 8B & 37.55\% & 35.26\% & 33.10\% & 5.43\% & 29.46\% & 18.15\% & 34.52\% & 15.71\% & 23.71\% \\
\textit{EduChat-R1-Qwen3-32B} & 32B & 37.25\% & 36.97\% & 43.33\% & 5.23\% & 38.93\% & 17.95\% & 39.31\% & 15.70\% & 25.92\% \\
\midrule
\multicolumn{11}{c}{\cellcolor{model_type}\textbf{Proprietary Models}} \\
\midrule
\textit{GPT-5.4} & -- & 58.36\% & 65.74\% & 37.90\% & 51.71\% & 37.45\% & 28.20\% & 48.26\% & 44.92\% & 46.34\% \\
\textit{GPT-5.6-Sol} & -- & 62.30\% & 68.59\% & 54.81\% & 54.39\% & 37.09\% & 32.80\% & 54.66\% & 48.26\% & 50.97\% \\
\textit{Claude-Opus-5} & -- & 50.87\% & 59.68\% & 67.64\% & 55.70\% & 55.27\% & 45.09\% & 56.42\% & 52.22\% & 54.01\% \\
\textit{GLM-5.3} & -- & 63.20\% & 58.46\% & 75.32\% & 10.90\% & 58.78\% & 22.55\% & \rankThree{65.40\%} & 22.39\% & 38.62\% \\
\textit{Kimi-K3} & -- & 69.18\% & 75.00\% & 66.09\% & 68.17\% & 52.03\% & 53.76\% & 64.06\% & \rankOne{63.39\%} & \rankOne{63.65\%} \\
\bottomrule
\end{tabular}}
\vspace{-2mm}
\end{table*}
\begin{table*}[t]
\centering
\caption{Detailed LongTutor results on evidence use, diagnosis, and teaching quality. Top-three values are shown in \rankOne{red}, \rankTwo{orange}, and \rankThree{yellow}, respectively.}
\label{tab:longtutor}
\resizebox{\linewidth}{!}{
\begin{tabular}{l c cccc cc cccccc}
\toprule
\multirow{2}{*}{\textbf{Model}} &
\multirow{2}{*}{\textbf{Size}} &
\multicolumn{4}{c}{\textsc{Evidence}} &
\multicolumn{2}{c}{\textsc{Diagnosis}} &
\multicolumn{6}{c}{\textsc{Teaching}} \\
\cmidrule(lr){3-6}
\cmidrule(lr){7-8}
\cmidrule(lr){9-14}
& & \textbf{IE} & \textbf{MR} & \textbf{HC} & \textbf{Avg.} &
\textbf{Acc} & \textbf{F1} &
\textbf{R-L} & \textbf{Hist.} & \textbf{Strat.} & \textbf{Coher.} & \textbf{Appr.} & \textbf{Avg.} \\
\midrule
\multicolumn{14}{c}{\cellcolor{model_type}\textbf{Open-weight Models}} \\
\midrule
\textit{Qwen3.5-4B-Base} & 4B & 49.84\% & 14.42\% & 12.76\% & 25.67\% & 37.89\% & 24.71\% & 0.2953 & 1.36 & 1.54 & 2.19 & 2.86 & 1.60 \\
\textit{\model-4B} (ours) & 4B & 80.84\% & 27.45\% & 89.34\% & 65.88\% & 51.51\% & 29.08\% & 0.3376 & 1.73 & 2.23 & 2.83 & 2.68 & 2.29 \\
\textit{Qwen3.5-9B-Base} & 9B & 10.12\% & 0.48\% & 6.84\% & 5.81\% & 23.71\% & 17.96\% & 0.3140 & 1.28 & 1.44 & 1.86 & 2.18 & 1.48 \\
\textit{\model-9B} (ours) & 9B & 84.61\% & 28.58\% & 86.71\% & 66.63\% & \rankTwo{53.55\%} & 26.13\% & 0.3469 & 2.04 & 2.60 & 3.18 & 3.00 & 2.66 \\
\textit{Qwen3.8-27B} & 27B & 34.31\% & 3.90\% & 72.17\% & 36.80\% & \rankThree{51.78\%} & \rankOne{38.41\%} & 0.3304 & 2.26 & 2.65 & 3.35 & 3.68 & \rankTwo{2.74} \\
\textit{\model-27B} (ours) & 27B & 92.87\% & 37.69\% & 68.24\% & \rankTwo{78.20\%} & \rankOne{54.04\%} & 30.58\% & 0.3457 & 2.56 & 2.98 & 3.50 & 3.95 & \rankOne{3.02} \\
\textit{Confucius3-Math} & 14B & 41.23\% & 11.95\% & 16.04\% & 23.07\% & 13.78\% & 10.16\% & 0.1653 & 1.25 & 1.34 & 1.72 & 1.98 & 1.36 \\
\textit{MuduoLLM} & 14B & 42.79\% & 13.94\% & 16.20\% & 24.31\% & 12.00\% & 6.46\% & 0.2352 & 1.19 & 1.27 & 1.65 & 2.09 & 1.28 \\
\textit{EduChat-SFT-Qwen2.5-7B} & 7B & 42.73\% & 5.65\% & 31.27\% & 26.55\% & 48.87\% & 19.69\% & 0.1028 & 1.14 & 1.18 & 1.66 & 2.05 & 1.21 \\
\textit{EduChat-R1-Qwen3-8B} & 8B & 64.91\% & 19.70\% & 16.63\% & 33.75\% & 18.41\% & 9.62\% & 0.0878 & 1.12 & 1.12 & 1.30 & 1.88 & 1.14 \\
\textit{EduChat-R1-Qwen3-32B} & 32B & 84.61\% & 26.48\% & 44.99\% & 52.03\% & 39.72\% & 24.47\% & 0.2480 & 1.44 & 1.50 & 2.09 & 2.82 & 1.56 \\
\midrule
\multicolumn{14}{c}{\cellcolor{model_type}\textbf{Proprietary Models}} \\
\midrule
\textit{GPT-5.4} & -- & 92.73\% & 59.89\% & 73.89\% & 75.50\% & 44.15\% & 30.94\% & 0.3525 & 1.37 & 1.58 & 2.08 & 1.98 & 1.62 \\
\textit{GPT-5.6-Sol} & -- & 94.05\% & 43.08\% & 95.35\% & \rankThree{77.50\%} & 45.09\% & \rankTwo{38.07\%} & 0.3651 & 1.59 & 1.79 & 2.31 & 2.21 & 1.89 \\
\textit{Claude-Opus-5} & -- & 96.33\% & 68.16\% & 88.03\% & \rankOne{84.17\%} & 28.57\% & 27.13\% & 0.2691 & 2.03 & 2.69 & 3.12 & 2.78 & \rankThree{2.72} \\
\textit{GLM-5.3} & -- & 94.09\% & 59.95\% & 78.23\% & 77.42\% & 40.59\% & \rankThree{34.50\%} & 0.3403 & 2.01 & 2.60 & 3.18 & 2.92 & 2.66 \\
\textit{Kimi-K3} & -- & 94.09\% & 52.96\% & 77.15\% & 74.73\% & 38.44\% & 30.97\% & 0.3520 & 1.75 & 2.07 & 2.63 & 2.52 & 2.17 \\
\bottomrule
\end{tabular}}
\vspace{-2mm}
\end{table*}

\subsection{General Capability Retention}
\label{app:general-capability}
In Tables~\ref{tab:ifeval}, \ref{tab:gpqa}, and \ref{tab:mmmu-pro}, we report the complete general-capability diagnostics used to measure whether education-oriented tuning changes instruction following, scientific reasoning, or multimodal reasoning. The paired base/SFT presentation makes the effect of tuning directly auditable.

\begin{table*}[t]
\centering
\scriptsize
\setlength{\tabcolsep}{4.5pt}
\caption{IFEval results for the base and education-tuned models. Prompt-level metrics require all verifiable instructions in a prompt to be satisfied, whereas instruction-level metrics average over individual instructions; strict and loose use the benchmark's two matching criteria~\cite{zhou2023ifeval}. For each model scale, the better result between the base and SFT variants is shown in \textbf{bold}.}
\label{tab:ifeval}
\begin{tabular}{l c cccc}
\toprule
\textbf{Model} & \textbf{Size} & \textbf{Prompt Strict} & \textbf{Inst. Strict} & \textbf{Prompt Loose} & \textbf{Inst. Loose} \\
\midrule
\textit{Qwen3.5-4B-Base} & 4B & 64.88\% & 73.50\% & 67.84\% & \textbf{76.50\%} \\
\textit{\model-4B} (ours) & 4B & \textbf{65.06\%} & \textbf{73.62\%} & \textbf{68.21\%} & 76.38\% \\
\textit{Qwen3.5-9B-Base} & 9B & 69.69\% & 78.18\% & 74.31\% & 81.41\% \\
\textit{\model-9B} (ours) & 9B & \textbf{73.01\%} & \textbf{80.34\%} & \textbf{75.05\%} & \textbf{81.77\%} \\
\textit{Qwen3.8-27B} & 27B & 80.59\% & 86.33\% & 84.84\% & \textbf{89.93\%} \\
\textit{\model-27B} (ours) & 27B & \textbf{82.07\%} & \textbf{87.89\%} & \textbf{85.40\%} & \textbf{89.93\%} \\
\bottomrule
\end{tabular}
\vspace{-2mm}
\end{table*}


\begin{table*}[t]
\centering
\setlength{\tabcolsep}{8pt}
\caption{GPQA accuracy on the Main and Diamond subsets for the base and education-tuned models~\cite{rein2023gpqa}. For each model scale, the better result between the base and SFT variants is shown in \textbf{bold}.}
\label{tab:gpqa}
\begin{tabular}{l ccc}
\toprule
\textbf{Model} & \textbf{Size} & \textbf{Main Acc.} & \textbf{Diamond Acc.} \\
\midrule
\textit{Qwen3.5-4B-Base} & 4B & 55.80\% & 58.08\% \\
\textit{\model-4B} (ours) & 4B & \textbf{56.03\%} & \textbf{60.61\%} \\
\textit{Qwen3.5-9B-Base} & 9B & 57.14\% & 62.12\% \\
\textit{\model-9B} (ours) & 9B & \textbf{61.61\%} & \textbf{63.64\%} \\
\textit{Qwen3.8-27B} & 27B & 70.76\% & 74.24\% \\
\textit{\model-27B} (ours) & 27B & \textbf{72.32\%} & \textbf{77.78\%} \\
\bottomrule
\end{tabular}
\vspace{-2mm}
\end{table*}


\begin{table*}[t]
\centering
\scriptsize
\setlength{\tabcolsep}{3.2pt}
\caption{MMMU-Pro accuracy across disciplines for the base and education-tuned models~\cite{yue2024mmmu_pro}. The abbreviated discipline headers correspond to Art and Design, Business, Science, Health and Medicine, Humanities, and Technology and Engineering. For each model scale, the better result between the base and SFT variants is shown in \textbf{bold}.}
\label{tab:mmmu-pro}
\begin{tabular}{l c rrrrrrr}
\toprule
\textbf{Model} & \textbf{Size} & \textbf{Overall} & \textbf{Art} & \textbf{Business} & \textbf{Science} & \textbf{Health} & \textbf{Humanities} & \textbf{Tech.} \\
\midrule
\textit{Qwen3.5-4B-Base}     & 4B  & 50.46\% & 59.21\% & 63.99\% & 49.14\% & 48.60\% & 56.75\% & 35.25\% \\
\textit{\model-4B} (ours)   & 4B  & \textbf{52.60\%} & 55.26\% & 64.34\% & 51.55\% & 53.50\% & 56.31\% & 41.25\% \\
\textit{Qwen3.5-9B-Base}     & 9B  & 58.38\% & 58.77\% & 76.92\% & 55.33\% & 56.99\% & 61.71\% & 46.76\% \\
\textit{\model-9B} (ours)   & 9B  & \textbf{60.75\%} & 56.58\% & 74.48\% & 61.51\% & 58.39\% & 66.21\% & 51.80\% \\
\textit{Qwen3.8-27B} & 27B & 64.97\% & 59.65\% & 82.52\% & 63.23\% & 62.59\% & 65.77\% & 58.27\% \\
\textit{\model-27B} (ours) & 27B & \textbf{67.98\%} & 60.09\% & 84.27\% & 69.07\% & 65.04\% & 67.12\% & 62.83\% \\
\bottomrule
\end{tabular}
\vspace{-2mm}
\end{table*}

\section{Training Configuration}
\label{app:training_details}

Table~\ref{tab:sft_config} summarizes the training configuration used for the \model models.
We perform full-parameter supervised fine-tuning with LLaMA-Factory.
The 4B, 9B and 27B models use the same training hyperparameters.
Unless otherwise specified, the remaining options follow the default settings of the training framework.

\begin{table}[t]
\centering
\small
\setlength{\tabcolsep}{5pt}
\caption{Training configuration for \model.}
\label{tab:sft_config}
\begin{tabular}{ll}
\toprule
\textbf{Component} & \textbf{Setting} \\
\midrule
Backbones &
Qwen3.5-4B-Base; Qwen3.5-9B-Base; Qwen3.8-27B \\
Framework &
LLaMA-Factory \\
Training Stage &
Supervised fine-tuning (SFT) \\
Fine-tuning Type &
Full-parameter fine-tuning \\
Template &
\texttt{qwen3\_5\_nothink} \\
Max Sequence Length &
32,768 \\
Packing &
False \\
Per-device Batch Size &
1 \\
Gradient Accumulation Steps &
8 \\
Learning Rate &
$5\times10^{-6}$ \\
LR Scheduler &
Cosine \\
Warmup Ratio &
0.1 \\
Epochs &
3 \\
Precision &
bf16 \\
Gradient Checkpointing &
Enabled \\
DeepSpeed &
ZeRO-3 \\
Preprocessing Workers &
1 \\
Dataloader Workers &
4 \\
\bottomrule
\end{tabular}
\end{table}

\end{document}